\documentclass[runningheads]{llncs}
\usepackage{graphicx}
\usepackage{comment}
\usepackage{amsmath,amssymb} 
\usepackage{color}
\usepackage{hyperref}

\usepackage{xspace}
\usepackage{mathtools,xparse}
\usepackage{xcolor}

\usepackage{subcaption}
\usepackage{amsmath,amsfonts,bm}

\def\eqref#1{equation~\ref{#1}}

\def\1{\bm{1}}

\newcommand{\test}{\mathcal{D_{\mathrm{test}}}}

\DeclareMathAlphabet{\mathsfit}{\encodingdefault}{\sfdefault}{m}{sl}
\SetMathAlphabet{\mathsfit}{bold}{\encodingdefault}{\sfdefault}{bx}{n}

\newcommand{\norm}[1]{\left\|#1\right\|} 

\makeatletter
\DeclareRobustCommand\onedot{\futurelet\@let@token\@onedot}
\def\@onedot{\ifx\@let@token.\else.\null\fi\xspace}

\def\eg{\emph{e.g}\onedot} 
\def\ie{\emph{i.e}\onedot} \def\Ie{\emph{I.e}\onedot}

\def\etal{\emph{et al}\onedot}
\makeatother

\newcommand\blfootnote[1]{%
  \begingroup
  \renewcommand\thefootnote{}\footnotetext{#1}%
  \addtocounter{footnote}{-1}%
  \endgroup
}

\begin{document}
\pagestyle{headings}
\mainmatter
\def\ECCVSubNumber{1422}  

\title{Why do These Match? Explaining the Behavior of Image Similarity Models} 

\titlerunning{Why do These Match?}
%
\author{Bryan A. Plummer*\inst{1} \and
Mariya I. Vasileva*\inst{2} \and
Vitali Petsiuk\inst{1} \and
Kate Saenko\inst{1,3} \and
David Forsyth\inst{2}
}
\authorrunning{B.\ A.\ Plummer et al.}
%
\institute{Boston University, Boston MA 02215, USA \and 
University of Illinois at Urbana-Champaign, Urbana IL 61801, USA \and
MIT-IBM Watson AI Lab, Cambridge MA 02142, USA\\
\email{\{bplum,vpetsiuk,saenko\}@bu.edu}\hspace{2mm} \email{\{mvasile2,daf\}@illinois.edu}} 
\maketitle
\begin{abstract}
Explaining a deep learning model can help users understand its behavior and allow researchers to discern its shortcomings. Recent work has primarily focused on explaining models for tasks like image classification or visual question answering.  In this paper, we introduce Salient Attributes for Network Explanation (SANE) to explain image similarity models, where a model's output is a score measuring the similarity of two inputs rather than a classification score.  In this task, an explanation depends on both of the input images, so standard methods do not apply. Our SANE explanations pairs a saliency map identifying important image regions with an attribute that best explains the match.  We find that our explanations provide additional information not typically captured by saliency maps alone, and can also improve performance on the classic task of attribute recognition. Our approach's ability to generalize is demonstrated on two datasets from diverse domains, Polyvore Outfits and Animals with Attributes 2. Code available at:
\url{https://github.com/VisionLearningGroup/SANE}
\keywords{Explainable AI, Image Similarity Models, Fashion Compatibility, Image Retrieval}
\end{abstract}

\blfootnote{*equal contribution}

\section{Introduction}
\begin{figure}[t]
    \centering
    \includegraphics[width=\textwidth,trim=0cm 3.9cm 0cm 0cm,clip]{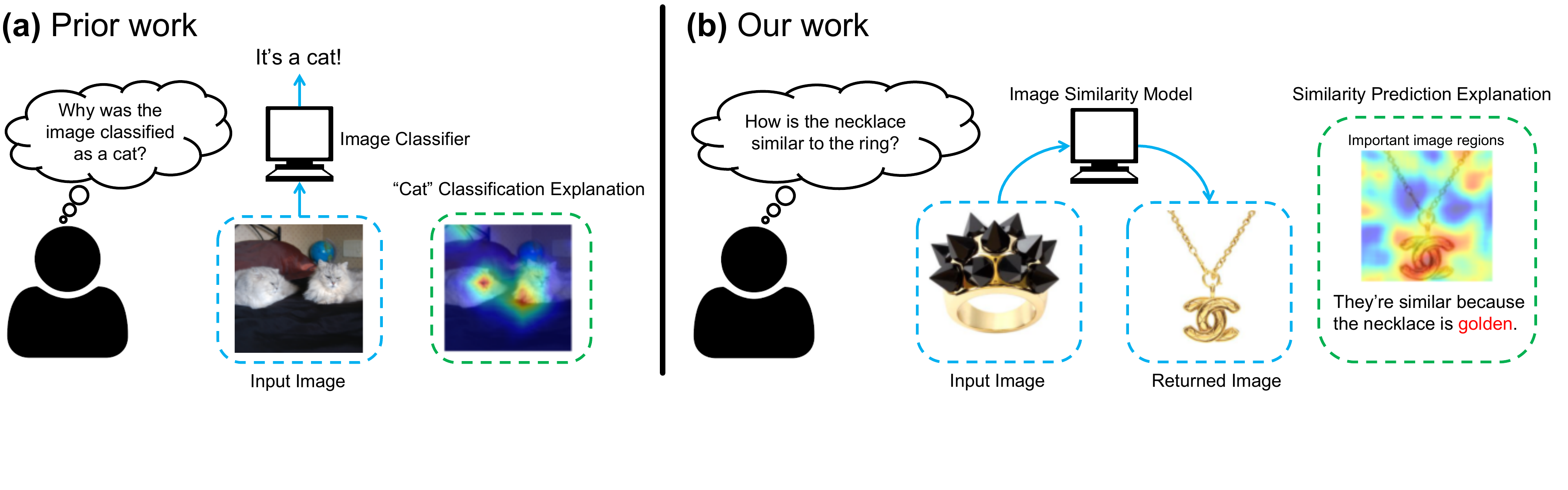}
    \caption{Existing explanation methods focus on image classification problems (left), whereas we explore explanations for image similarity models (right). We pair a saliency map, which identifies important image regions but often provides little interpretable information, with an attribute (\eg, golden), which is more human-interpretable and, thus, a more useful explanation than saliency alone}
    \label{fig:motivational_example}
\end{figure}

Many problems in artificial intelligence that require reasoning about complex relationships can be solved by learning a feature embedding to measure similarity between images and/or other modalities such as text.  Examples of these tasks include image retrieval~\cite{kiapourICCV15where,Radenovi_2018_CVPR,Yelamarthi_2018_ECCV}, zero-shot recognition~\cite{Bansal_2018_ECCV,Li_2018_CVPR,Wang_2018_CVPR} or scoring fashion compatibility~\cite{hanACMMM2017,Hsiao_2018_CVPR,tanSimilarity2019,VasilevaECCV18FasionCompatibility}.  Reasoning about the behavior of similarity models can aid researchers in identifying potential improvements, show where two images differ for anomaly detection, promote diversity in fashion recommendation by ensuring different traits are most prominent in the top results, or simply help users understand the model's predictions which can build trust~\cite{teachtrust}.    However, prior work on producing explanations for neural networks has primarily focused on explaining classification models~\cite{fong_iccv_2017,nguyen2016synthesizing,Petsiuk2018rise,lime:kdd16,Selvaraju_2017_ICCV,zeilerECCV2014} and does not directly apply to similarity models. Given a \emph{single} input image, such methods  produce a saliency map which identifies pixels that played a significant role towards a particular class prediction (see Figure~\ref{fig:motivational_example}a for an example). 
On the other hand, a similarity model requires at least \emph{two} images to produce a score. 
The interaction between both images defines which features are more important, so replacing just one of the images can result in identifying different salient traits.  

Another limitation of existing work is that saliency alone may be insufficient as an explanation of (dis)similarity. When similarity is determined by the presence or absence of an object, a saliency map 
may be enough to understand model behavior. However, for the image pair in Figure~\ref{fig:motivational_example}b, 
highlighting the necklace as the region that contributes most to the similarity score is reasonable, but uninformative given that there are no other objects in the image.
Instead, what is important is that the necklace shares a similar color with the ring. Whether these
attributes or salient parts
are a better fit can vary depending on the image pairs and the domain they come from. For example, an image can be matched as formal-wear because of a
shirt's collar (salient part), while two images of animals can match because both have stripes (attribute).

Guided by this intuition, we introduce \emph{Salient Attributes for Network Explanation (SANE)}. Our approach generates a saliency map to explain a model's similarity score, paired with an attribute explanation that identifies important image properties.  SANE is a ``black box'' method, meaning it can explain any network architecture
and only needs to measure changes to a similarity score with different inputs. Unlike a standard classifier, which simply predicts the most likely attributes for a given image, our explanation method predicts which attributes are important for the similarity score predicted by a model. Predictions are made for each image in a pair, and allowed to be non-symmetric: \eg, the explanation for why the ring in Figure~\ref{fig:motivational_example}b matches the necklace may be that it contains ``black'', even though the explanation for why the necklace matches the ring could be that it is ``golden.'' A different similarity model may also result in different attributes being deemed important for the same pair of images. 

 SANE combines three major components: an attribute predictor, a prior on the suitability of each attribute as an explanation, and a saliency map generator. Our underlying assumption is that at least one of the attributes present in each image should be able to explain the similarity score assigned to the pair. Given an input image, the attribute predictor provides a confidence score and activation map for each attribute, while the saliency map generator produces regions important for the match. During training, SANE encourages overlap between the similarity saliency and attribute activation. At test time, we rank attributes as explanations for an image pair based on a weighted sum of this attribute-saliency map matching score, the explanation suitability prior of the attribute, and the likelihood that the attribute is present in the image. Although we only evaluate the top-ranked attribute, in practice multiple attributes could be used to explain a similarity score. We find that using saliency maps as supervision for the attribute activation maps during training not only improves the attribute-saliency matching, resulting in  better attribute explanations, but also boosts attribute recognition performance using standard metrics like average precision.

We evaluate several candidate saliency map generation methods which are primarily adaptations of ``black box'' approaches that do not rely on a particular model architecture or require access to network parameters to produce a saliency map~\cite{fong_iccv_2017,Petsiuk2018rise,lime:kdd16,zeilerECCV2014}. These methods generally identify important regions by measuring a change in the output class score resulting from a perturbation of the input image. Similarity models, however, often rely on a learned embedding space to reason about relationships between images, where proximity between points or the lack thereof indicates some degree of correspondence. An explanation system for embedding models must therefore consider how distances between embedded points, and thus their similarity, change based on perturbing one or both input images. We explore two strategies for adapting these approaches to our task. First, we manipulate just a single image (the one we wish to produce an explanation for) while keeping the other image fixed. Second, we manipulate both images to allow for more complex interactions between the pair. See Section~\ref{sec:saliency_maps} for details and a discussion on the ramifications of this choice.

Our paper makes the following contributions: 1) we provide the the first quantitative study of explaining the behavior of image similarity models; 2) we propose a novel explanation approach that combines saliency maps and attributes; 3) we validate our method with a user study combined with metrics designed to link our explanations to model performance, and find that it produces more informative explanations than adaptations of prior work to this task, and further improves attribute recognition performance.


\section{Related Work}
\begin{figure}[t]
    \centering
    \includegraphics[width=\textwidth, trim=1.7cm 2cm 2.3cm 0cm,clip]{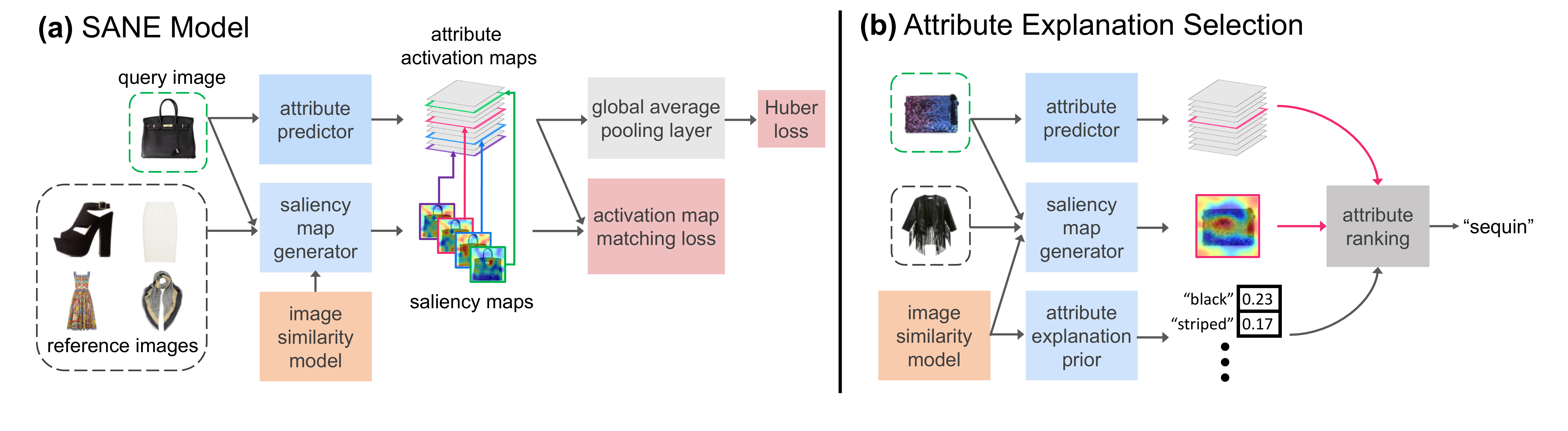}
    \caption{
    \textbf{Approach Overview.} (a) During training we use the saliency map generator (Section~\ref{sec:saliency_maps}) to the important regions when compared to many reference images.  Then, we encourage at least one ground truth attribute's activation maps to match each saliency map (details in Section~\ref{sec:attributes}). (b) At test time, we rank attribute explanations by how well the saliency and attribute activation maps match, along with the likelihood of the attribute and its explanation suitability prior (details in Section~\ref{sec:inference}). We assume the image similarity model has been pretrained and is kept fixed in all our experiments}
    \label{fig:attribute_model}
\end{figure}

\noindent\textbf{Saliency-based Explanations.}  
Saliency methods can generally be split into ``white box'' and ``black box'' approaches.  
``White box'' methods assume access to internal components of a neural network, either in the form of gradients or activations of specific layers~\cite{Cao_2015_ICCV,changICLR2019,nguyen2016synthesizing,Selvaraju_2017_ICCV,simonyanICLR2014,yosinski-2015-ICML-DL-understanding-neural-networks,zhang2016EB,zhou2015cnnlocalization}.  
Most of these methods produce a saliency map by using some version of backpropagation from class probability to an input image.
In contrast, ``black box'' methods require no knowledge of model internals (\eg weights or gradients).
They obtain saliency maps by perturbing the input in a predefined way and measuring the effect of that perturbation on the model output, such as class score.
We adapt and compare three ``black box'' and one ``white box'' methods for our saliency map generator in Figure~\ref{fig:attribute_model}.  
``Black box'' approaches include a Sliding Window~\cite{zeilerECCV2014}, which masks image regions sequentially, and Randomized Input Sampling for Explanations (RISE)~\cite{Petsiuk2018rise}, which masks random sets of regions. Both measure the effect removing these regions has on the class score. 
LIME~\cite{lime:kdd16} first obtains a super-pixel representation of an image.  Super-pixel regions are randomly deleted, and their importance is estimated using Lasso.  
``White box'' Mask~\cite{fong_iccv_2017} learns a saliency map directly by using different perturbation operators and propagating the error to a low resolution mask.  Although there exists limited work that adapts certain saliency methods to the image similarity setting~\cite{Gordo_2017_CVPR}, they present qualitative results only, \ie, these methods are not evaluated quantitatively on their explanation accuracy as done in our work.
\smallskip

\noindent\textbf{Natural Language and Attribute-based Explanations.}  Instead of producing saliency maps, which can sometimes be difficult to interpret, researchers have explored methods of producing text-based explanations.  These include methods which justify a model's answer in the visual question answering task~\cite{Park_2018_CVPR,Li_2018_ECCV}, rationalize the behavior of a self-driving vehicle~\cite{Kim_2018_ECCV}, or describe why a category was selected in fine-grained object classification~\cite{hendricksECCV2016}. Lad~\etal~\cite{Lad2014} used human-generated attribute explanations describing why two images are similar or dissimilar as guidance for image clustering.  Our approach could be used to automatically generate these explanations rather than relying on human feedback. Several works exist which learn attribute explanations either to identify important concepts~\cite{Bau_2017_CVPR,Fong_2018_CVPR} or to justify a model's decision by pointing to evidence~\cite{Hendricks_2018_ECCV}. Kim~\etal~\cite{kimTCAVICML2018} learns a concept activation vector that separates examples with an attribute against examples without it, then scores the sensitivity of attributes based on how often a directional derivative changes the inputs towards the concept.
However, all these methods were designed to explain categorical predictions rather than similarity models.  To the best of our knowledge, ours is the first work which uses attribute explanations in the image similarity setting.
\smallskip

\noindent\textbf{Interpretable Image Similarity.}  Attributes are often used to provide a sense of interpretability to for image similarity tasks~\cite{HanInterpretable2019,liaoACMMM2018,yang2019interpretable} or aid in the retrieval of similar images using attribute information~\cite{AkCVPR2018,Zhao_2017_CVPR}.  However, these methods typically require that a model be trained with a particular architecture in order to provide an interpretabile output (\ie, they cannot be directly applied to any pretrained model).  In contrast, SANE is able to explain the predictions of any image similarity model regardless of architecture.

\section{Salient Attributes for Network Explanations (SANE)}

We are given a fixed model that predicts the similarity between two images, and must explain why a query image is similar to a reference image. While typical models for predicting similarity are learned from data, \eg, with an embedding method and triplet loss, our approach is agnostic as to how the model being explained is built. SANE consists of three components: the attribute explanation model (Section~\ref{sec:attributes}), the saliency map generator (Section~\ref{sec:saliency_maps}), and an attribute explanation suitability prior (Section~\ref{sec:inference}).  Although we train a CNN to produce attribute predictions, the image similarity model we wish to explain is kept fixed. At test time, one recovers a saliency map for the match from the query image in a pair, then uses the attribute explanation model and attribute suitability prior to rank each attribute's ability to explain the image similarity model.  See Figure~\ref{fig:attribute_model} for an overview of our approach.

\subsection{Attribute Explanation Model}
\label{sec:attributes}


Suppose we have access to pairs of images $(I_r, I_q)$, where $I_r$ denotes a reference image, and $I_q$ a query image. We wish to obtain an explanation for the match between $I_r$ and $I_q$. Associated with each pair is a saliency map $\mathbf{m}_q$ produced by a saliency map generator (described in Section~\ref{sec:saliency_maps}). To compute a saliency map for $\mathbf{m}_r$ instead, we need simply to swap the query and reference images, which would likely result in a different saliency map than $\mathbf{m}_q$. Finally, assume we have access to binary attribute annotations $a_i$, $i = 1, \ldots, A$, and let $\mathbf{a}_{gt} \in \{0, 1\}^A$ be the set of ground truth attribute annotations for a given query image.  If no attribute annotations are provided, an attribute discovery method could be employed (\eg,~\cite{han2017automatic,VittayakornECCV2016}).  We explore an attribute discovery method in the appendix.

Our attribute explanation model produces confidence scores $\hat{\mathbf{a}} \in \mathbb{R}^A$ for $I_q$. Unlike a standard attribute classifier, however, our goal is not just to predict the most likely attributes in $I_q$, but rather to identify which attributes contribute the most to the similarity score $s(I_r, I_q)$ produced by the  similarity model we wish to obtain explanations for. To accomplish this, the layer activations for attribute $a_i$ before the global average pooling layer are defined as an attribute activation map $\mathbf{n}_i$.  This attribute activation map represents a downsampled mask of an image that identifies prominent regions in $I_q$ for that attribute.  We encourage at least one ground truth attribute's activation map for image $I_q$ to match saliency map $\mathbf{m}_q$ as a form of regularization. Our underlying assumption, which we validate empirically, is that at least one of the ground truth attributes of $I_q$ should be able to explain why $I_q$ is similar to $I_r$. Thus, at least one of the attribute activation maps $\mathbf{n}_i$ should closely resemble the saliency map for the match, $\mathbf{m}_q$.

Each attribute confidence score is obtained using a global average pooling layer on its attribute activation map, followed by a softmax. A traditional loss function for multi-label classification would be binary cross-entropy, which makes independent (\ie, noncompetitive) predictions reflecting the likelihood of each attribute in an image.  However, this typically results in a model where attribute scores are not comparable.  For example, a confidence score of 0.6 may be great for attribute $A$, but a horrible score for attribute $B$.  Thus, such a loss function would be ill-suited for our purposes since we need a ranked list of attributes for each image. Instead, our attribute explanation model is trained using a Huber loss~\cite{huberloss}, sometimes referred to as a smooth $\ell_1$ loss, which helps encourage sparsity in predictions. This provides a competitive loss across attributes and thus can help ensure calibrated attribute confidence scores that can be used to rank attribute prevalence in an image. More formally, given a set of confidence scores $\hat{\mathbf{a}}$ and attribute labels $\mathbf{a}_{gt}$, our loss is,
\begin{equation}
    L_{Huber}(\hat{\mathbf{a}}, \mathbf{a}_{gt}) = 
    \begin{cases}
    \frac{1}{2}(\mathbf{a}_{gt} - \hat{\mathbf{a}})^2 & \text{for } |\mathbf{a}_{gt} - \hat{\mathbf{a}}| \leq 1\\
    |\mathbf{a}_{gt} - \hat{\mathbf{a}}| & \text{otherwise.}
    \end{cases}
\end{equation}
Note that multiple attributes can be present in the image; and that this loss operates on attributes, not attribute activation maps. Since the confidence scores sum to one (due to the softmax), we scale a binary label vector by the number of ground truth attributes $A_{gt}$ (\eg, if there are four attributes for an image, its label would be 0.25 for each ground truth attribute, and zero for all others).
\smallskip

\noindent\textbf{Leveraging saliency maps during training.} We explicitly encourage  our model to identify attributes that are useful in explaining the predictions of an image similarity model by finding which attributes best describe the regions of high importance to similarity predictions. 
To accomplish this, we first find a set of regions that may be important to the decisions of an image similarity model by generating a set of $K$ saliency maps $\mathcal{M}_q$ for up to $K$ reference images similar to the query. For the query image under consideration, we also construct a set of attribute activation maps $\mathcal{N}_{gt}$ corresponding to each ground truth attribute. Then, for each saliency map in $\mathcal{M}_q$, we find its best match in $\mathcal{N}_{gt}$.  We match saliency maps to attributes rather than the other way around since not all annotated attributes are necessarily relevant to the explanation of $s(I_r, I_q)$. We use an $\ell_2$ loss between the selected attribute activation map and saliency map, \ie,
\begin{equation}
     L_{hm} = \frac{1}{K}\sum_{\forall \mathbf{m} \in \mathcal{M}_q} \min_{\forall \mathbf{n} \in  \mathcal{N}_{gt}} \norm{\mathbf{m} - \mathbf{n}}_{2}.
     \label{eq:heatmap_loss}
\end{equation}
Combined with the attribute classification loss, our model's complete loss is:
\begin{equation}
    L_{total} = L_{Huber} + \lambda L_{hm},
    \label{eq:attr_loss}
\end{equation}
where $\lambda$ is a scalar parameter.

\subsection{Saliency Map Generator}
\label{sec:saliency_maps}

Most ``black box'' methods produce a saliency map by measuring the effect manipulating the input image (\eg, by removing image regions) has on a model's similarity score. If a large drop in similarity is measured, then the region must be significant.  If almost no change is measured, then the model considers the image region irrelevant. The saliency map is generated by averaging the similarity scores for each pixel over all instances where it was altered.  The challenge is determining the best way of manipulating the input image to discover these important regions.  A key benefit of ``black box'' methods is that they do not require having access to underlying model parameters. We compare three black box methods: a simple Sliding Window baseline~\cite{zeilerECCV2014}, LIME~\cite{lime:kdd16}, which determines how much super-pixel regions affect the model predictions, and RISE~\cite{Petsiuk2018rise}, an efficient high-performing method that constructs a saliency map using random masking.  We also compare to ``white box'' learned Mask~\cite{fong_iccv_2017}, which was selected due to its high performance and tendency to produce compact saliency maps.
We now describe how we adapt these models for our task; see appendix for additional details on each method.
\smallskip

\noindent{\bf Computing similarity scores}. Each saliency method we compare is designed to operate on a single image, and measures the effect manipulating the image has on the prediction of a specific object class. However, an image similarity model's predictions are arrived at using two or more input images. Let us consider the case 
where we are comparing only two images -- a query image (\ie the image we want to produce an explanation for) and a single reference image, although our approach extends to consider multiple reference images. Even though we do not have access to a class label, we can measure the effect manipulating an image has on the similarity score between the query and reference images.  Two approaches are possible:  manipulating both images, or only the query image.

{\bf Manipulating both images} would result in $NM$ forward passes through the image similarity model (for $N$ query and $M$ reference image manipulations), which is prohibitively expensive unless $M<< N$. But we only need an accurate saliency map for the query image, so we set $M << N$ in our experiments.  There is another danger: for example, consider two images of clothing items that are similar if either they both contain or do not contain a special button. Masking out the button in one image and not the other would cause a drop in similarity score, but masking out the button in both images would result in high image similarity. These conflicting results could make accurately identifying the correct image regions contributing to a score difficult.

The alternative is to {\bf manipulate the query image} alone, and use a fixed reference.  We evaluate saliency maps produced by both methods in Section~\ref{sec:saliency_eval}.



\subsection{Selecting Informative Attributes}
\label{sec:inference}

At test time, given a similarity model and a pair of inputs, SANE generates a saliency map and selects an attribute to show to the user.  We suspect that some attributes are not useful for explaining a given image similarity model. Thus, we take into account each attribute's usefulness by learning concept activation vectors (CAVs)~\cite{kimTCAVICML2018} over the final image similarity embedding.  These CAVs identify which attributes are useful in explaining a layer's activations by looking at whether an attribute positively affects the model's predictions.  CAVs are defined as the vectors that are orthogonal to the classification boundary of a linear classifier trained to recognize an attribute over a layer's activations.  Then, the sensitivity of each concept to an image similarity model's predictions (the TCAV score) is obtained by finding the fraction of features that are positively influenced by the concept using directional derivatives computed via triplet loss with a margin of machine epsilon. Note that this creates a single attribute ranking over the entire image similarity embedding (\ie, it is agnostic to the image pair being explained), which we use as an attribute explanation suitability prior.
Finally, attributes are ranked as explanations using a weighted combination of the TCAV scores, the attribute confidence score $\hat{\mathbf{a}}$, and how well the attribute activation map $\mathbf{n}$ matches the generated saliency map $\mathbf{m}_q$. \Ie,
\begin{equation}
    \mathbf{e}(\mathbf{m}_q, \hat{\mathbf{a}}, \mathbf{n}, \text{TCAV}) =  \phi_1 \hat{\mathbf{a}} + \phi_2 \; d_{\text{cos}}(\mathbf{m}_q, \mathbf{n}) + \phi_3 \text{TCAV},
    \label{eq:attr_score}
\end{equation}
where $d_{\text{cos}}$ denotes cosine similarity, and $\phi_{1-3}$ are scalars estimated via grid search on held out data. See the appendix for additional details.
\section{Experiments}

\textbf{Datasets.} We evaluate our approach using two datasets from different domains to demonstrate its ability to generalize.  The Polyvore Outfits dataset~\cite{VasilevaECCV18FasionCompatibility} consists of 365,054 fashion product images annotated with 205 attributes and composed into 53,306/10,000/5,000 train/test/validation outfits.  Animals with Attributes 2 (AwA)~\cite{awa2dataset} consists of 37,322 natural images of 50 animal classes annotated with 85 attributes, and is split into 40 animal classes for training, and 10 used at test time. To evaluate our explanations, we randomly sample 20,000 (query, reference) pairs of images for each dataset from the test set, where 50\% of the pairs are annotated as similar images.
\smallskip

\noindent\textbf{Image Similarity Models.}  For Polyvore Outfits we use the type-aware embedding model released by Vasileva~\etal~\cite{VasilevaECCV18FasionCompatibility}.  This model captures item compatibility (\ie how well two pieces of clothing go together) using a set of learned projections on top of a general embedding, each of which compares a specific pair of item types (\ie a different projection is used when comparing a top-bottom pair than when comparing a top-shoe pair).  For AwA we train a feature representation using a 18-layer ResNet~\cite{He_2016_CVPR} with a triplet loss function that encourages animals of the same type to embed nearby each other. For each dataset/model, cosine similarity is used to compare an image pair's feature representations.

\subsection{Saliency Map Evaluation}
\label{sec:saliency_eval}

\textbf{Metrics.}  Following Petsiuk~\etal~\cite{Petsiuk2018rise}, we evaluate the generated saliency maps using insertion and deletion metrics which measure the change in performance of the model being explained as pixels are inserted into a blank image, or deleted from the original image. For our task, we generate saliency maps for all query images, and insert or delete pixels in that image only. If a saliency map correctly captures the most important image regions, we should expect a sharp drop in performance as pixels are deleted (or a sharp increase as they are inserted). We report the area under the curve (AUC) created as we insert/delete pixels at a rate of 1\% per step for both metrics. We normalize the similarity scores for each image pair across these thresholds so they fall in a [0-1] interval.
\smallskip

\noindent\textbf{Results.}  Table~\ref{tab:saliency_results} compares the different saliency map generation methods on the insertion and deletion tasks. RISE performs best on most metrics, with the exception of LIME doing better on the deletion metric on AwA. This is not surprising, since LIME learns which super-pixels contribute to a similarity score. For AwA this means that parts of the animals could be segmented out and deleted or inserted in their entirety before moving onto the next super-pixel. On Polyvore Outfits, however, the important components may be along the boundaries of objects (\eg the cut of a dress), something not well represented by super-pixel segmentation. Although Mask does not perform as well as other approaches, it tends to produce the most compact regions of salient pixels as it searches for a saliency map with minimal support (see the appendix for examples). Notably, we generally obtain better performance when the reference image is kept fixed and only the query image is manipulated. This may be due to issues stemming from noisy similarity scores as discussed in Section~\ref{sec:saliency_maps}, and suggests extra care must be taken when manipulating both images.

\begin{table}[t]
\centering
\setlength{\tabcolsep}{3pt}
    \caption{Comparison of candidate saliency map generator methods described in Section~\ref{sec:saliency_maps}.  We report AUC for the insertion and deletion metrics described in Section~\ref{sec:saliency_eval}}
    \begin{tabular}{lccccc}
    \hline
     & Fixed & \multicolumn{2}{c}{Polyvore Outfits} & \multicolumn{2}{c}{Animals with Attributes 2}\\
    Method &  Reference? & Insertion ($\uparrow$) & Deletion ($\downarrow$) & Insertion ($\uparrow$) & Deletion ($\downarrow$)\\
    \hline
    \hline
        Sliding Window & Y & 57.1 & 50.6 & 76.9 & 76.9\\
        LIME & Y & 55.6 & 52.7 & 76.9 & \textbf{71.7}\\
        Mask & Y & 56.1 & 51.8 & 72.4 & 75.9\\
        RISE & Y & \textbf{61.2} & \textbf{46.8} & \textbf{77.8} & 74.9\\
        \hline
        Sliding Window & N & 56.6 & 51.1 & 77.6 & 76.5\\
        Mask & N & 55.6 & 52.6 & 72.9 & 76.6\\
        RISE & N & 58.5 & 50.6 & 77.7 & 73.8\\
        \hline
    \end{tabular}
    \label{tab:saliency_results}
\end{table}


\subsection{Attribute Prediction Evaluation}
\label{sec:attr_pred_eval}

\textbf{Metrics.}  For the standard task of attribute recognition we use mean average precision (mAP) computed over predictions of all images in the test set.  Two additional metrics are used to evaluate our attribute explanations using the (query, reference) image pairs from the saliency map experiments, which are similar to the evaluation of saliency maps.  Given the set of attributes we know exist in the image, we select which attribute among them best explains the similarity score using Eq.~(\ref{eq:attr_score}), and then see the effect \emph{deleting} that attribute from the image has on the similarity score.  Analogically, we select the attribute which best explains the similarity score from those which are \emph{not} present in the image, and measure the effect \emph{inserting} that attribute has on the similarity score.  Intuitively, if an attribute is critical for an explanation, the similarity score should shift more than if a different attribute is selected.  Scores for these metrics are expressed in terms of relative change.  When inserting important missing attributes, we expect that the similarity score to improve, and vice versa: when deleting important attributes, we would expect the similarity score to drop. 

\begin{figure}[t]
    \centering
    \includegraphics[width=0.7\textwidth,trim=0cm 11.2cm 8.2cm 0cm,clip]{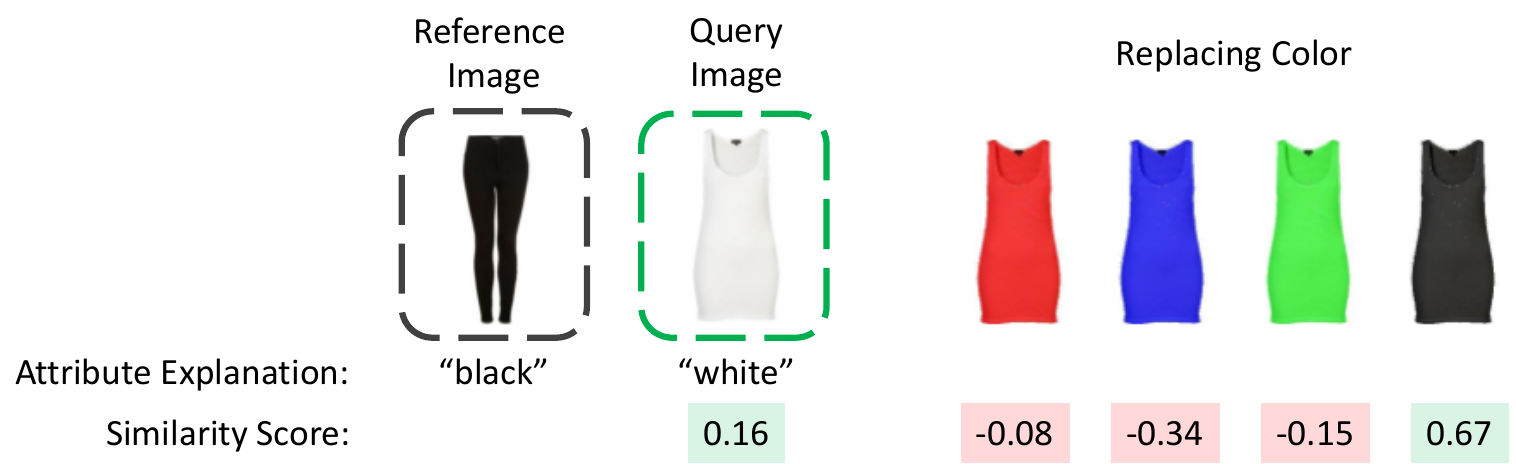}
    \caption{\textbf{Attribute replacement example.} First, SANE explains why the similarity model predicted that the leggings (reference) and  dress (query) have a compatibility score 0.16 -- because the dress is "white" and leggings are "black." Then, we artificially change the color of the dress and re-compute similarity.  Since our explanation was a good one, this lowers compatibility most of the time.  However, this can be noisy as compatibility did improve for the black dress, but still useful as we show in Section~\ref{sec:human_study}}
    \label{fig:color_replacement}
\end{figure}

We provide an example of attribute removal and its effect on similarity in Figure~\ref{fig:color_replacement}.  Note that, because the attribute explanation is a color in this example, we can easily remove the attribute by replacing with another color. We see that when we modify the white dress to be a different color, the similarity score drops significantly.  The only exception is when we make the dress the same color (black) as the attribute explanation of the pants it is being compared to.  This demonstrates in a causal way how our predicted explanation attributes can play a significant role in the similarity scores. 

Since most attributes are not as easily replaced as colors, in order to insert or remove a particular attribute, we find a \emph{representative image} in the test set which is most similar to the query image in all other attributes.  For example, let us consider a case where we want to remove the attribute ``striped'' from a query image.  We would search through the database for images which are most similar in terms of non-striped attributes, but which have not been labeled as being ``striped''.  We rank images using average confidence for each attribute computed over the three attribute models we compare in Table~\ref{tab:attr_results}.  After obtaining these average attribute confidence scores, we use cosine similarity between the non-explanatory attributes to score candidate representative images.  On the Polyvore Outfits dataset we restrict the images considered to be of the same type as the query image.  For example, if the query image is a shoe, then only a shoe can be retrieved.
After retrieving the representative image, we compute its similarity with the reference image using the similarity model to compare with the original (query, reference) pair. Examples of this process can be found in the appendix.  Since any retrieved image may inadvertently change multiple attributes, we average the scores over the top-k representative images.  
\smallskip

\noindent\textbf{Compared methods.}  We provide three baseline approaches: a random baseline, a simple attribute classifier (\ie no attribute activation maps), and a modified version of FashionSearchNet~\cite{AkCVPR2018} -- an attribute recognition model which also creates a weakly-supervised attribute activation map, for comparison. To validate our model choices, we also compare using a binary cross-entropy loss $L_{BCE}$ to the Huber loss for training our attribute predictors.  Additional details on these models can be found in the appendix.
\smallskip

\noindent\textbf{Results.}  Table~\ref{tab:attr_results} shows the performance of the compared attribute models for our metrics. Our attribute explanation metrics demonstrate the effectiveness of our attribute explanations, with our model, which matches saliency maps and includes TCAV scores, obtaining best performance on both datasets. This shows that ``inserting'' or ``deleting'' the attribute predicted by SANE from the query image affects the similarity model's predicted score more than inserting or deleting the attribute suggested by baselines. Notably, our approach outperforms FashionSearchNet + Map Matching (MM), which can be considered a weakly-supervised version of SANE trained for attribute recognition.  The fifth line of Table~\ref{tab:attr_results} reports that TCAV consistently outperforms many methods on insertion, but has mixed results on deletion.  This is partly due to the fact that other models, including SANE, are trained to reason about attributes that actually exist in an image, whereas for insertion the goal is to predict which attribute \emph{not} present in the image would affect the similarity score most significantly.  Thus, using a bias towards globally informative attributes (\ie, TCAV scores) is more consistently useful for insertion.  Finally, training an attribute classifier using saliency maps for supervision ($L_{hm}$) leads to a 3\% improvement on the standard attribute recognition task measured in mAP over a simple attribute classifier while using the same number of parameters.  This also outperforms FashionSearchNet, which treats localizing important image regions as a latent variable rather than using saliency maps for supervision.  While $L_{BCE}$ does perform well for attribute recognition, it does poorly as an explanation as shown in Table~\ref{tab:attr_results}.


\begin{table}[t]
\centering
\setlength{\tabcolsep}{1.pt}
    \caption{Comparison of how attribute recognition (mAP) and attribute explanation (insertion, deletion) metrics described in Section~\ref{sec:attr_pred_eval} are affected for different approaches. We use fixed-reference RISE as our saliency map generator for both datasets}
    \begin{tabular}{lcccccc}
    \hline
    & \multicolumn{3}{c}{Polyvore Outfits} & \multicolumn{3}{c}{Animals with Attributes 2} \\
    \hline
    Method & mAP & Insert ($\uparrow$) & Delete ($\downarrow$) & mAP & Insert ($\uparrow$) & Delete ($\downarrow$)\\
    \hline
    \hline
    Random & -- & 25.3 & -6.3  & -- & 2.1 & -8.5\\
    \hline
    Attribute Classifier - $L_{BCE}$ & 53.2 & 25.7 & -5.8 & 65.1 & -2.5 & -2.3 \\
    Attribute Classifier - $L_{BCE} + L_{hm}$ & \textbf{56.0} & 25.8 & -5.9 & \textbf{67.2} & -1.7 & -2.0 \\
    FashionSearchNet - $L_{BCE}$~\cite{AkCVPR2018} & 54.7 & 25.5 & -5.2 & 65.9 & -1.6 & -2.6\\
    \hline
    TCAV~\cite{kimTCAVICML2018} & -- & 28.0 & -8.5 & -- & 3.4 & -22.0\\
    Attribute Classifier - $L_{Huber}$ & -- & 25.9 & -8.1 & -- & -0.8 & -2.8 \\
    FashionSearchNet - $L_{Huber}$~\cite{AkCVPR2018} & -- & 26.1 & -7.6 & -- & -0.3 & -3.5\\
    FashionSearchNet - $L_{Huber}$ + MM & -- & 26.6 & -10.1 & -- & 1.2 & -6.8\\
    SANE & -- & 26.3 & -9.8 & -- & 0.1 & -3.0\\
    SANE + MM & -- & 27.1 & -10.9 & -- & 6.0 & -10.7\\
    SANE + MM + TCAV (Full) & -- & \textbf{31.5} & \textbf{-11.8} & -- & \textbf{6.2} & \textbf{-24.1}\\
    \hline
    \end{tabular}
    \label{tab:attr_results}
\end{table}

\begin{figure}[t]
    \centering
    \includegraphics[width=\textwidth,trim=0cm 0cm 0cm 0cm,clip]{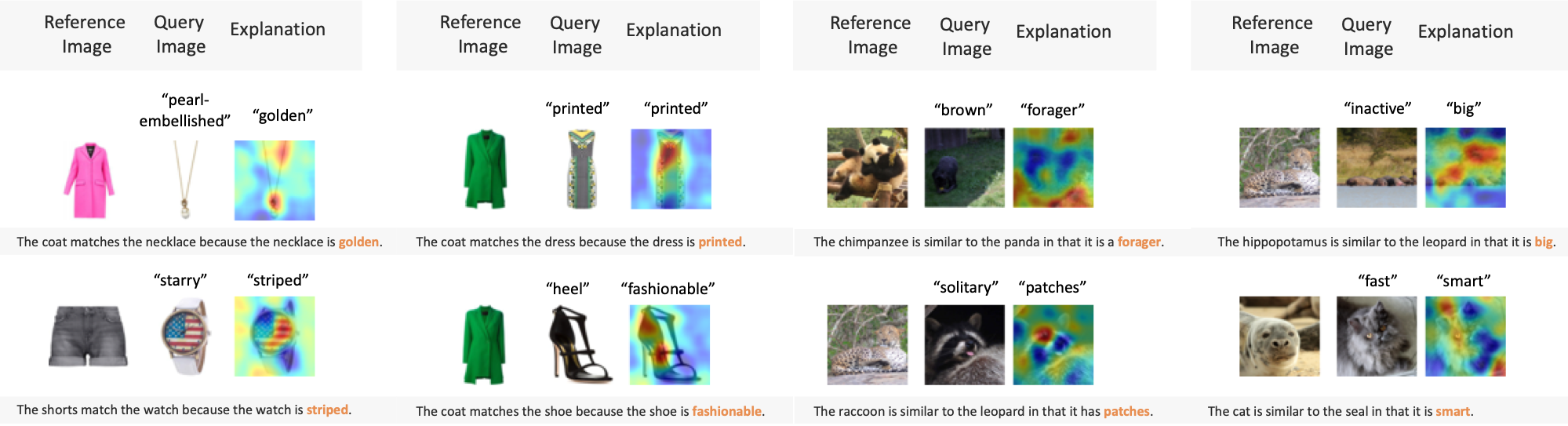}
    \caption{Qualitative results of our attribute explanations for pairs of examples on the Polyvore Outfits and the AwA datasets. The attribute predicted as explanation for each reference-query match is shown above the saliency map. The most likely attribute for the query image as predicted by our attribute classifier is shown directly above it}
    \label{fig:attribute_examples}
\end{figure}

We provide qualitative examples of our explanations in Figure~\ref{fig:attribute_examples}. Examples demonstrate that our explanations pass important sanity checks. 
Notice that ``golden'', ``striped'' and ``printed'' in the first two columns of Figure~\ref{fig:attribute_examples} are sensibly localized, and are also reasonable explanations for the match, while a more abstract explanation like ``fashionable'' is linked to the high heel, the curve of the sole, and the straps of the shoe. Note further that the explanations are non-trivial: they more often than not differ from the most likely attribute in the query image, as predicted by a standard attribute classifier. In other words, our explanation model is utilizing information from each pair of images and the saliency map characterizing the match to produce a sensible, interpretable explanation. 


\subsection{User Study}
\label{sec:human_study}
A key component of evaluating whether our explanations are sensible and interpretable is conducting a user study. This is not a straightforward task, as one has to carefully formulate the prompts asked in a manner suitable for answering questions like, \emph{Are our explanations useful? Do they provide insight that helps users understand what the similarity model is doing? Are they consistent?} 

We formulate our study as a ``guessing game" whereby a unique image triplet $(A, B, C)$ is presented in each question that asks users to select whether the image similarity model predicted B or C as a better match for $A$. Images $B$ and $C$ are selected such that the pairs $(A, B)$ and $(A, C)$ have sufficiently different similarity scores. We adopt an A/B testing methodology and present seven different versions of the study: (a) a control case whereby no explanations are presented and users have to guess whether $(A, B)$ or $(A, C)$ are more similar based on their own intuition; (b) image pairs $(A, B)$ and $(A, C)$ are presented along with a random saliency map for images $B$ and $C$; (c) image pairs $(A, B)$ and $(A, C)$ are presented along with the corresponding saliency maps for $B$ and $C$ generated using RISE~\cite{Petsiuk2018rise}; (d) $(A, B)$ and $(A, C)$ are presented along with a sentence explanation containing the most likely predicted attribute made by our attribute predictor for $B$ and $C$; (e) image pairs $(A, B)$ and $(A, C)$ are presented along with a sentence explanation containing the SANE explanation attribute for each pair; (f) image pairs $(A, B)$ and $(A, C)$ are presented along with \emph{both} randomly generated saliency maps for $B$ and $C$ and the SANE attributes for each pair;  and (g) image pairs $(A, B)$ and $(A, C)$ are presented along with \emph{both} the corresponding RISE saliency maps for $B$ and $C$ and the SANE attributes for each pair. Examples of each question type are provided in the appendix.

Our underlying hypothesis is that, if our model produces sensible and human-interpretable explanations, there must be a correlation between the resulting explanations and the similarity score for an image pair, and that relationship should be discernible by a human user. Thus, user accuracy on guessing which pair the model predicted is a better match based on the provided explanations or lack thereof should be a good indicator as to the reliability of our explanations.

Using a web form, we present a total of 50 unique image triplets to 59 subjects in the age range 14-50, each triplet forming a question as described above. We keep question order the same for each subject pool and question type. Each user sees 10 unique questions, and subject pools across different study versions are kept disjoint. Each question type received 1-4 unique responses.
\smallskip

\noindent\textbf{Study Results.} Table~\ref{tab:user_study} reports the results of our study. In addition to reporting the user's ability to correctly identify the image pair the similarity model thought was a better match, we also report the percent of subjects for each study type that reported finding the provided explanations helpful in answering the questions. Comparing lines 1 and 2 of Table~\ref{tab:user_study}, we see that, as expected, users find random saliency maps to be confusing, as 20\% or fewer ever report them as useful.  Comparing lines 1 and 3 of Table~\ref{tab:user_study}, we see that accuracy goes down on the Polyvore Outfits dataset when users are shown saliency maps, suggesting that users may have misinterpreted why the particular image regions were highlighted; yet all users thought the maps were helpful in answering the questions. However, on AwA, accuracy increases significantly with providing saliency in addition to image pairs, along with $87.5\%$ of users finding them helpful. 

\begin{table}[t]
    \caption{User study results. We report user's accuracy given different information in guessing which image pair the image similarity model thought was a better match and the portion of users who felt the explanations were helpful} 
\centering
    \begin{tabular}{rlcccc}
    \hline
    && \multicolumn{2}{c}{Polyvore Outfits} & \multicolumn{2}{c}{Animals with Attributes 2}\\
    \hline
    &Explanation type & Accuracy & Helpful? & Accuracy & Helpful? \\
    \hline
    \hline
        (a) & Control Case (no explanations) & 66.0 & -- & 42.0 & --\\
        \hline
        (b) & Random Maps & 66.0 & 20.0 & 50.0 & 18.2\\
        (c) & Saliency Maps & 56.7 & \bf{100.0} & 53.8 & \bf{87.5}\\
        \hline
        (d) & Predicted Attr's & 62.5 & 66.7 & 53.1 & 69.2\\
        (e) & SANE Attr's & 68.9 & 66.7 & \bf{61.3} & \bf{87.5}\\
        \hline
        (f) & Random Maps + SANE Attr's & 65.8 & 75.0 & 59.2 & 66.7\\
        (g) & Saliency Maps + SANE Attr's & \bf{70.0} & 55.6 & 52.5 & 62.5\\
    \hline
    \end{tabular}
    \label{tab:user_study}
\end{table}

Comparing the most likely attribute vs.\ our SANE explanation attributes in lines 4 and 5 of Table~\ref{tab:user_study}, respectively, we see that users demonstrate improved understanding of the image similarity model's behavior and also find the attribute explanations to be helpful.  On Polyvore Outfits, although users consider the most likely attributes as explanations about as helpful as SANE explanations, the $6.4\%$ difference in accuracy suggests that SANE explanations do indeed provide valuable information about the match. On AwA, SANE attributes result in the highest user accuracy, with $87.5\%$ of users reporting them useful vs.\ only $69.2\%$ reporting most likely attributes used as explanations useful. Notably, SANE attributes reports improving user accuracy by 3-7.5\% on both datasets over the control case or using saliency maps alone. Line 7 of Table~\ref{tab:user_study} shows that on Polyvore Outfits, users do best if they are provided both saliency maps and SANE attributes as explanations, even though they did not find this type of explanation most useful.  We suspect this could be due to natural human bias, \ie users' intuition disagreeing with the image similarity model's predictions.

Overall, the study results suggest that (1) users find our explanations helpful; (2) our explanations consistently outperform baselines; (3) the type of explanation that proves most helpful depends on the dataset: users like having saliency maps as a guide, although their performance is best using both saliency maps and explanation attributes on a fashion dataset, while on a natural image dataset, having attribute explanations for each image pair helps the most.



\section{Conclusion}

In this paper we introduced SANE, a method of explaining an image similarity model's behavior by identifying attributes that are important to the similarity score paired with saliency maps indicating significant image regions.  We confirm that our SANE explanations improve a person's understanding of a similarity model's behavior through a user study to supplement automatic metrics.  In future work, we believe closely integrating the saliency generator and attribute explanation model, enabling each component to take advantage of the predictions of the other, would help improve performance.
\smallskip

\noindent\textbf{Acknowledgements.}  This work is funded in part by a DARPA XAI grant, NSF Grant No. 1718221, and ONR MURI Award N00014-16-1-2007.
%
%
\bibliographystyle{splncs04}
\bibliography{egbib}

\appendix

\section{Candidate Salience Map Generator Descriptions}

\begin{figure}
    \centering
    \includegraphics[height=0.93\textheight,trim=0cm 0cm 0cm 0cm,clip]{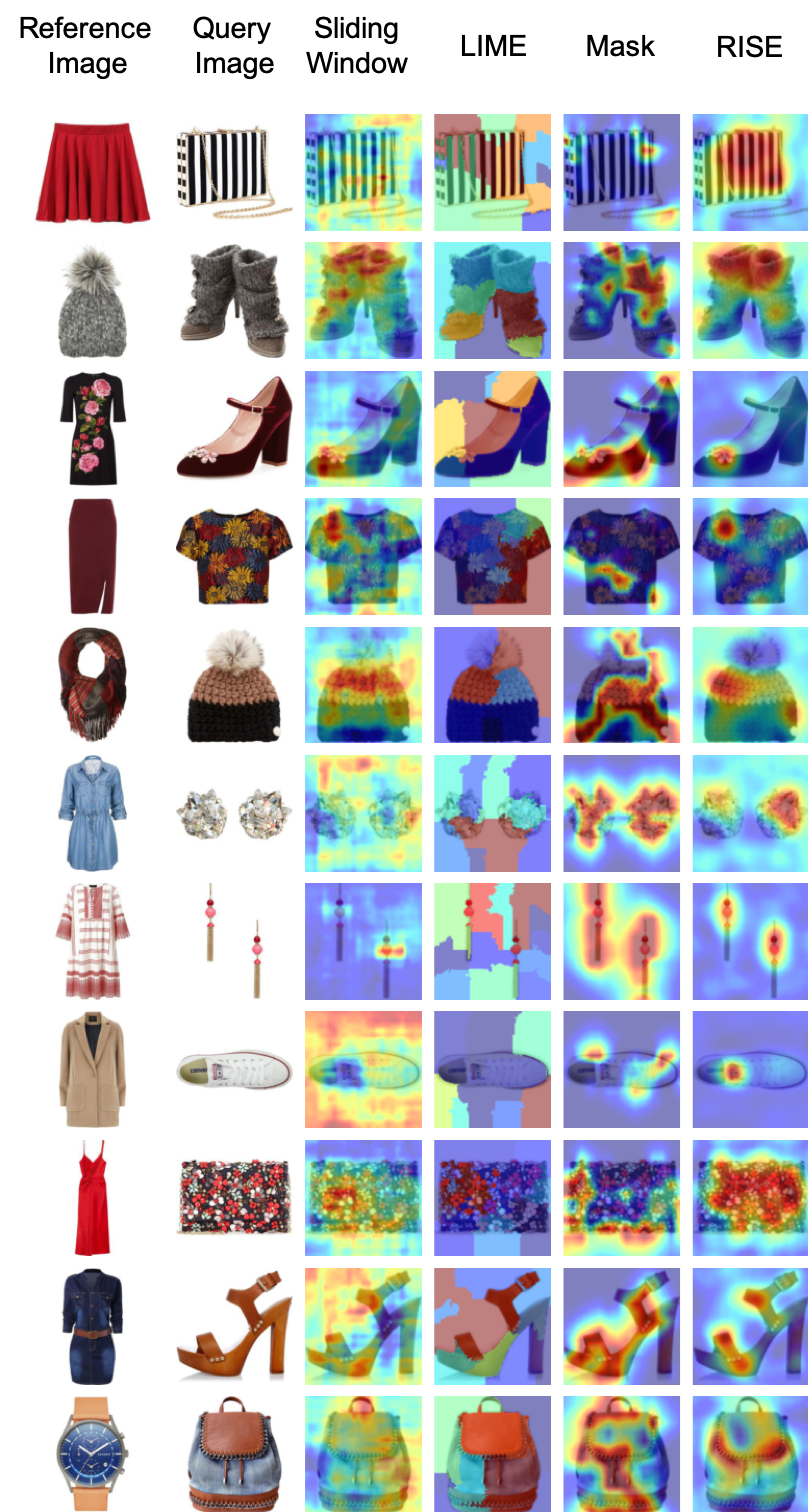}
    \caption{Qualitative examples comparing the saliency map generator candidates on the Polyvore Outfits dataset.}
    \label{fig:sup1_examples}
\end{figure}

\begin{figure}
    \centering
    \includegraphics[height=0.93\textheight,trim=0cm 0cm 0cm 0cm,clip]{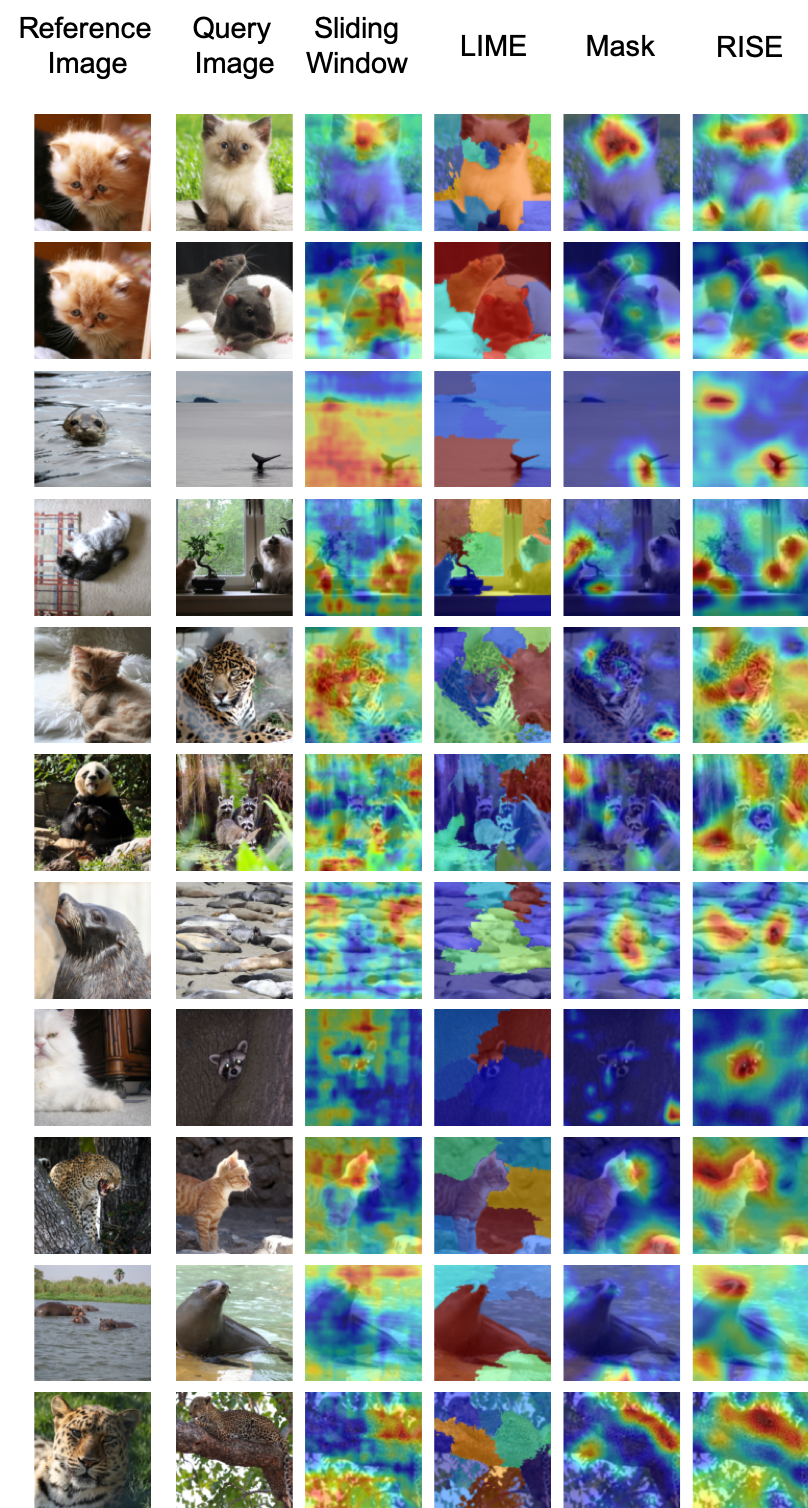}
    \caption{Qualitative examples comparing the saliency map generator candidates on the Animals with Attributes dataset.}
    \label{fig:sup2_examples}
\end{figure}

In this section we provide additional details about each of the candidate saliency map generation methods used in our paper. We split these approaches into two groups: methods which analyze behavior solely through input manipulation (described in Section~\ref{sec:saliency_manipulation}) and those which use an optimization procedure to learn some parameters in combination with input manipulation (described in Section~\ref{sec:salience_learned}).  Please see Section 3.2 of our paper for a description of how these methods are adapted to our task.  We also provide a runtime comparison of each approach in Table~\ref{tab:runtime}. A qualitative comparison between saliency map generators on the Polyvore Outfits and AwA datasets is provided in Figures~\ref{fig:sup1_examples} and~\ref{fig:sup2_examples}.

\subsection{Saliency Maps by Input Manipulation}
\label{sec:saliency_manipulation}
A straightforward approach to producing a saliency map is to manipulate the input image by removing image regions and measuring the effect this has on the similarity score.  If a large drop in similarity is measured, then the region must be important to this decision.  If almost no change was measured, then the model considers the image region irrelevant. The saliency map is generated from this approach by averaging the similarity scores for each pixel location over all instances where it was removed from the input.  The challenge then is to determine how to manipulate the input image to discover these important regions.  
\smallskip

\noindent\textbf{Sliding Window~\cite{zeilerECCV2014}.} The first approach to removing regions of an image we shall discuss is a sliding window, where regions are sampled regularly across an image.  There is a direct tradeoff, however, with how densely frames are sampled and the computational time it takes to do a forward pass through the network for each manipulated image.  If frames are not densely sampled to enable an efficient solution, then it wouldn't be able to localize important regions accurately. If regions are too densely sampled then removing them might not make enough of a difference in the similarity score to take measurements accurately.  When manipulating the inputs of the reference image, we apply 625 occlusion windows each covering a square region of about 12\% of image area. When manipulating both images we apply 36 occlusion windows to the reference image.
\smallskip

\noindent\textbf{RISE~\cite{Petsiuk2018rise}.} This method uses Monte Carlo approach to generate saliency maps.  A set of $N$ random binary masks of size $h \times w$ is sampled where each element is independently set to 1 with probably $p$, and all other elements are set to 0.  
Typically these masks are much smaller than the input image, so they are upsampled using bilinear interpolation.  This produces small continuous regions within the upsampled mask that can be used to manipulate the input image. 
To remove the fixed grid structure the masks are upsampled to larger than image size and then cropped randomly.
For both datasets we randomly sample 2,000 random masks upsampled from $8 \times 8$ mask with the probability of preserving a region of $0.5$. When manipulating the inputs of the reference image, we generate 30 random masks. Although this approach does require a significant number of random masks, we found this approach significantly outperforms using a sliding window that samples a similar number of masks on our task.

\subsection{Learned Saliency Maps}
\label{sec:salience_learned}

We shall now discuss methods which combine input manipulation with an optimization procedure used to directly learn a saliency map.  As in Section~\ref{sec:saliency_manipulation}, we compare generating saliency maps for a single query image at a time using a fixed reference image as well as generating a saliency map by manipulating both the query and reference images.
\smallskip

\noindent\textbf{LIME~\cite{lime:kdd16}.} Rather than masking regions without any concern over the continuity of a region, this approach to generating saliency maps operates over a superpixel segmentation of an image.  Images are manipulated by randomly deleting superpixels in the image.  After sampling $N=1000$ manipulated inputs, the importance of each superpixel is estimated using Lasso.   Finally, important regions are selected using submodular optimization.
\smallskip

\noindent\textbf{Mask~\cite{fong_iccv_2017}.}  In this approach a low resolution saliency map is directly learned using stochastic gradient decent and upsampled to the image size.  Instead of manipulating an image by just deleting regions as in other methods, two additional perturbation operators are defined: adding Gaussian noise and image blurring.  To help avoid artifacts when learning the mask a total-variation norm is used in addition to an $L1$ regularization to promote sparsity. This approach removes the reliance on superpixels and tends to converge in fewer iterations than LIME, although it is considerably slower in practice than other approaches (see Table~\ref{tab:runtime}).  That said - one advantage it does have over other approaches is the ability to learn the salience map for both the query and reference image jointly (which we take advantage of when we are not using a fixed reference image).  We learn a $14 \times 14$ perturbation mask for both datasets.  We train the mask for 500 iterations using Adam~\cite{adam} with a learning rate of 0.1.

\begin{table}[t]
    \centering
    \caption{Runtime comparison of the compared saliency generation methods and how using a fixed reference image, or manipulating both the query and reference images affects performance.}
    \begin{tabular}{lcc}
        \hline
        Method & Fixed Reference? & Time(s)\\
        \hline
        Sliding Window & Y & 0.2\\
        LIME & Y & 1.2\\
        Mask & Y & 4.1\\
        RISE & Y & 0.3\\
        \hline
        Sliding Window & N & 2.5\\
        Mask & N & 7.2\\
        RISE & N & 5.8\\
        \hline
    \end{tabular}
    \label{tab:runtime}
\end{table}

\newpage 
\section{Additional Experimental and Implementation Details}

\subsection{Compared Attribute Methods}
In this section we describe the attribute prediction methods we use as baselines for our attribute experiments in Section 4.2.  More details on our method are provided in Section~\ref{sec:sane}.  For our Attribute Classifier, FashionSearchNet, and SANE predictor we use the same  50-layer ResNet base image encoder~\cite{He_2016_CVPR} and the last convoluntional layer of the network the same number of channels as the number of classes.  The output of the last convolutional layer we refer to as our Attribute Activation Maps in our paper, and we use a global average pooling (GAP) layer to obtain attribute scores from these Attribute Activation Maps.

\begin{itemize}
    \item \textbf{Random baseline.} For insertion we randomly select an attribute not in the image to insert into the image.  Analogously, for deletion we randomly select an attribute that is in the image to be removed.
    \item \textbf{Attribute Classifier.}  This model uses a our image encoder as described earlier as a simple baseline.  Although classic methods typically use a fully connected layer rather than a convolutional layer followed by a GAP layer, we found the latter (\ie, our approach) to improve performance 2-3 mAP in our experiments.  This model has the same architecture as our SANE attribute predictor, but unlike the SANE model, it doesn't get any kind of supervision of its Attribute Activation Map.
    \item \textbf{FashionSearchNet~\cite{AkCVPR2018}.} This network uses an Attribute Activation Map to identify and extract a region of interest for each attribute.  These extracted regions are fed into two branches consisting of three fully connected layers which is trained for both attribute classification and image retrieval.  We remove the image retrieval components in our experiments.  This model provides a weakly-supervised baseline of an Attribute Activation Map to compare to SANE's saliency-supervised activation map.
\end{itemize}

\subsection{SANE Details}
\label{sec:sane}

\begin{figure}
    \centering
    \includegraphics[height=0.94\textheight,trim=0cm 0cm 0cm 0cm,clip]{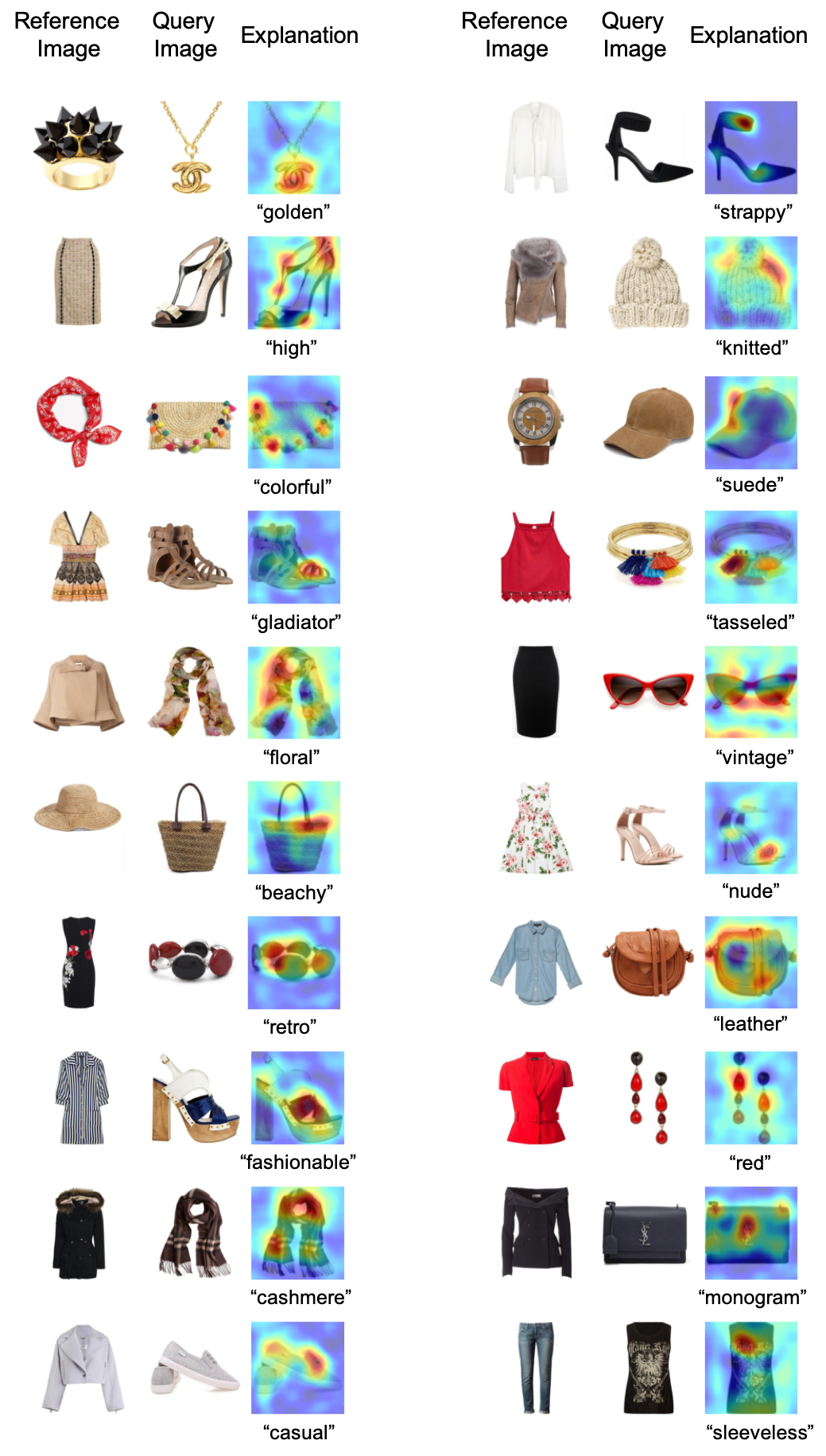}
    \caption{Additional qualitative examples of our SANE explanations on the Polyvore Outfits dataset.}
    \label{fig:sup2_3_examples}
\end{figure}

\begin{figure}
    \centering
    \includegraphics[height=0.94\textheight,trim=0cm 0cm 0cm 0cm,clip]{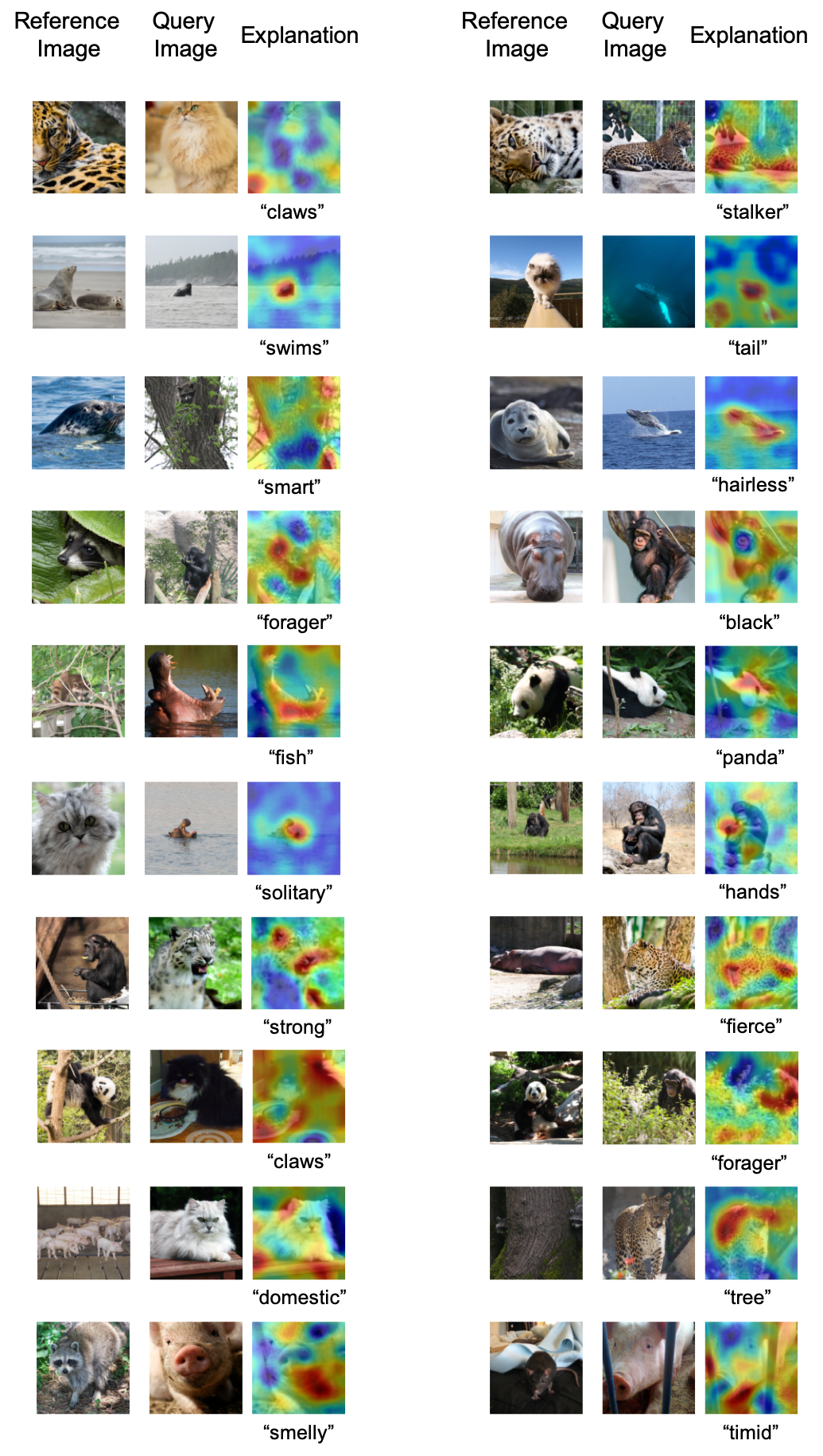}
    \caption{Additional qualitative examples of our SANE explanations on the AwA dataset.}
    \label{fig:sup4_5_examples}
\end{figure}

Due to its efficient (see Table~\ref{tab:runtime}) and overall good performance (see Table 1) we selected the fixed-reference RISE as our saliency map generator.  For each training image, we sample up to five similar images using the ground truth annotations of each dataset and generate saliency maps using each sampled image as the reference image.  We train our attribute model for 300 epochs using Adam~\cite{adam} with a learning rate of $5e^{-4}$ and set $\lambda = 5e^{-3}$ in Eq. 3 from the paper.  After each epoch, we computed mAP on the validation set and kept the best performing model according to this metric. Additional qualitative examples for explanations produced by our SANE model on the Polyvore Outfits and AwA datasets are provided in Figures~\ref{fig:sup2_3_examples} and~\ref{fig:sup4_5_examples}. 

\begin{figure}
    \centering
    \includegraphics[width=\textwidth,trim=0cm 4cm 0cm 0cm,clip]{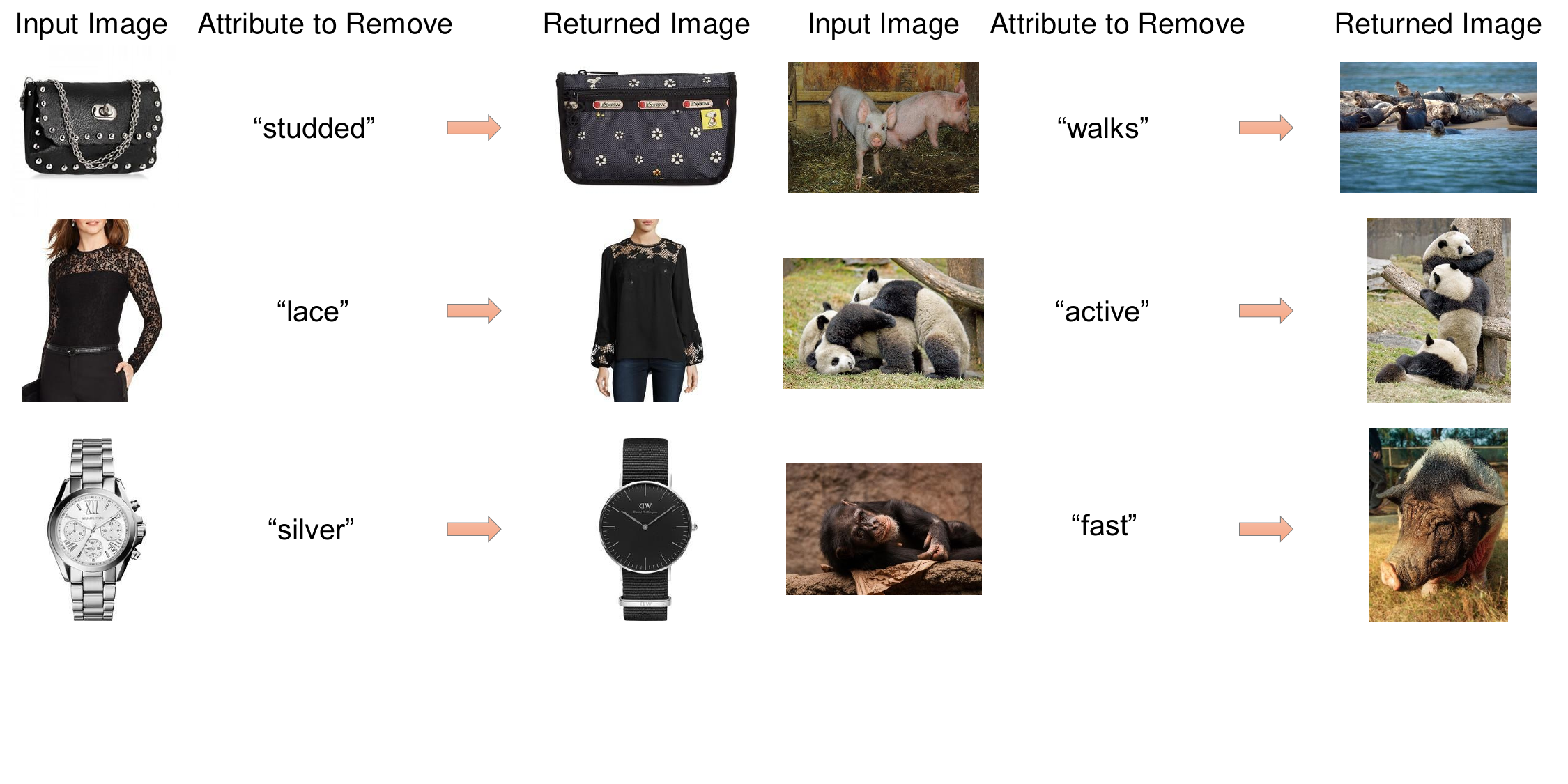}
    \caption{Examples of the attribute deletion process used to evaluate how good an attribute is as an explanation.  We measure the similarity of the input image and some reference image as well as between the returned image and the reference image.  If a large change in similarity is measured then the attribute is considered a ``good'' explanation.  If similarity stays about the same, the attribute is considered a ``poor'' explanation, \eg, trying to remove ``active'' from the pandas on the right.}
    \label{fig:removal_example}
\end{figure}

We provide an example of the attribute deletion process in Figure~\ref{fig:removal_example}.  After identifying an attribute to remove in an image, we search for the $K$ most similar image to the input from a database that doesn't contain the input attribute.  Image similarity is computed over the attribute space, \ie, we want to keep the predictions of each attribute the same, and only vary the target attribute.  For Polyvore Outfits, we only consider images of the same type (\ie, so tops can only be replaced with other tops).  To ensure we don't bias towards a single attribute model, average the predictions made by each attribute model in our experiments (Attribute Classifier, FashionSearchNet, and SANE).

We see on the left side of Figure~\ref{fig:removal_example} that some attributes like colors are largely retained when the attribute has to do with a non-color based attribute.  On the returned AwA images on the right side of Figure~\ref{fig:removal_example} we see how some attributes can lead to significant changes in the images or almost none at all depending on the attribute selected to remove.  For some items we can see that multiple attributes may have changed in the retrieved images in these figures.  To try to account for this, we compute our attribute insertion/deletion metrics over the $K=5$ most similar images to the input returned by our image selection procedure and then average the change in similarity using these 5 images.  While this does provide a noisy estimate of the change in similarity, our human evaluation in Section 4.4 shows that there is a correlation between gains in our automatic metrics and improvements in a human user's understanding of predictions made by an image similarity model.

\begin{figure}
    \centering
    \includegraphics[width=\textwidth]{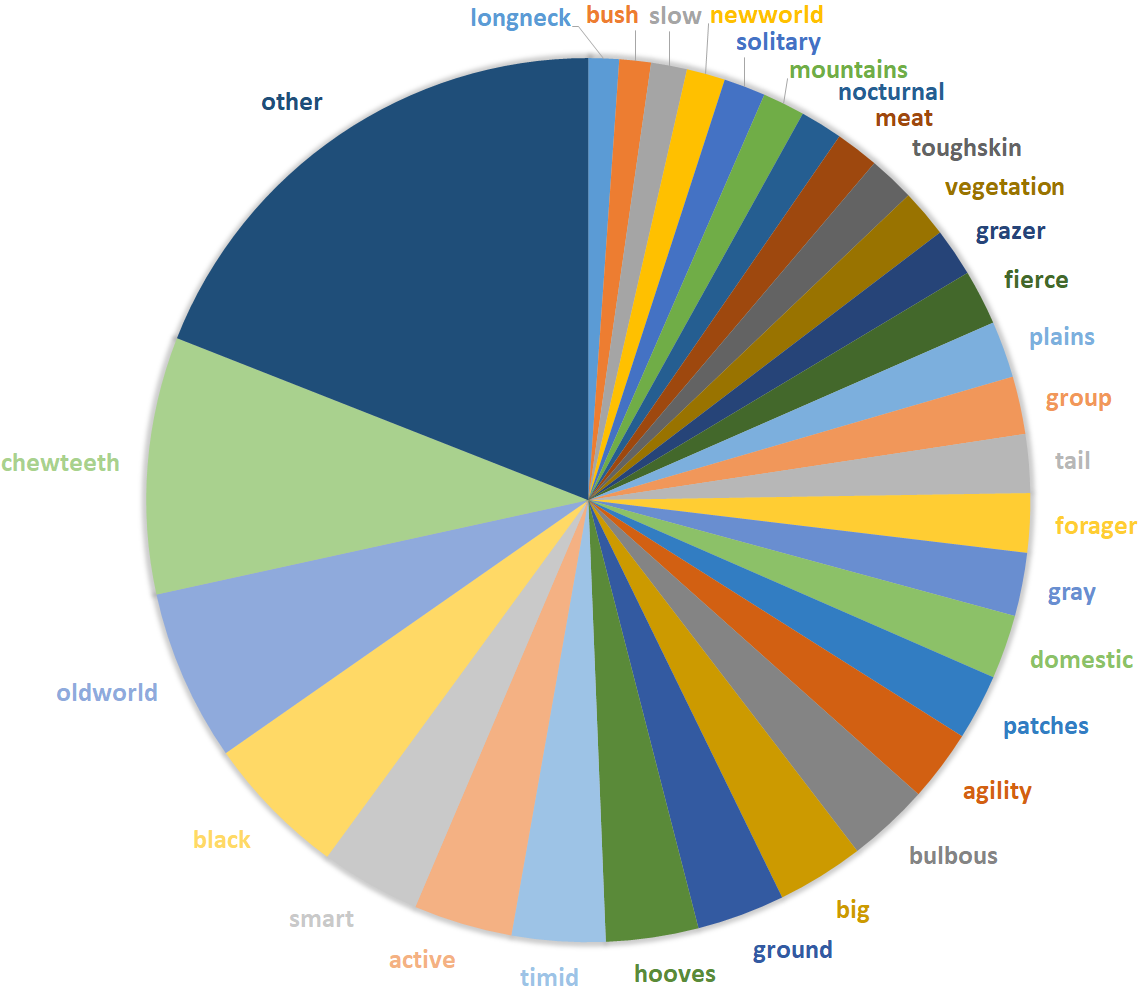}
    \caption{The likelihood each attribute in the AwA dataset was identified as the best attribute for an image pair on held-out data.}
    \label{fig:awa_prior}
\end{figure}

In Section 3.3 we discuss how we estimate how likely each attribute is a ``good'' explanation in held-out data.  This is used as a prior to bias our attribute selections towards attributes that are known to be good attribute explanations.  In Figure~\ref{fig:awa_prior} we show the ground truth bias for the attribute detection task according to our metrics for the AwA dataset.   Note, however, that this prior would change for a different image similarity model.  For example, if the image similarity model was more biased towards colors, then we would expect to see the likelihood for ``black,'' ``brown,'' and ``gray'' to increase.



\section{User Study}
\label{sec:supp_human_study}

\begin{figure}[t]
    \centering
    \includegraphics[width=\textwidth,trim=0.1cm 0cm 0cm 0cm,clip]{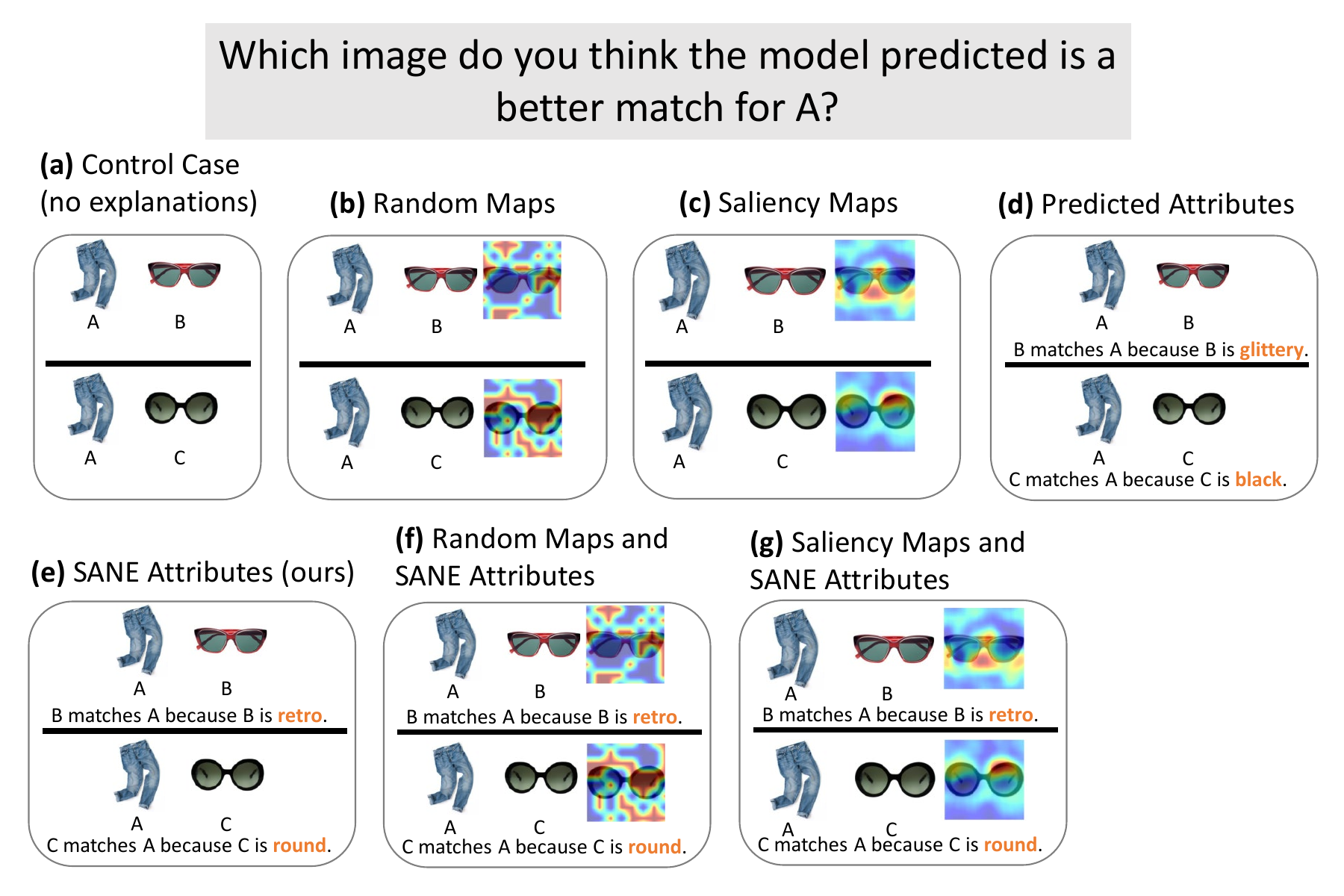}
    \caption{Variants of our user study for a given image triplet on the Polyvore Outfits dataset.}
    \label{fig:supp_user_study}
\end{figure}

\begin{figure}[t]
    \centering
    \includegraphics[width=\textwidth,trim=0.1cm 0cm 0cm 0cm,clip]{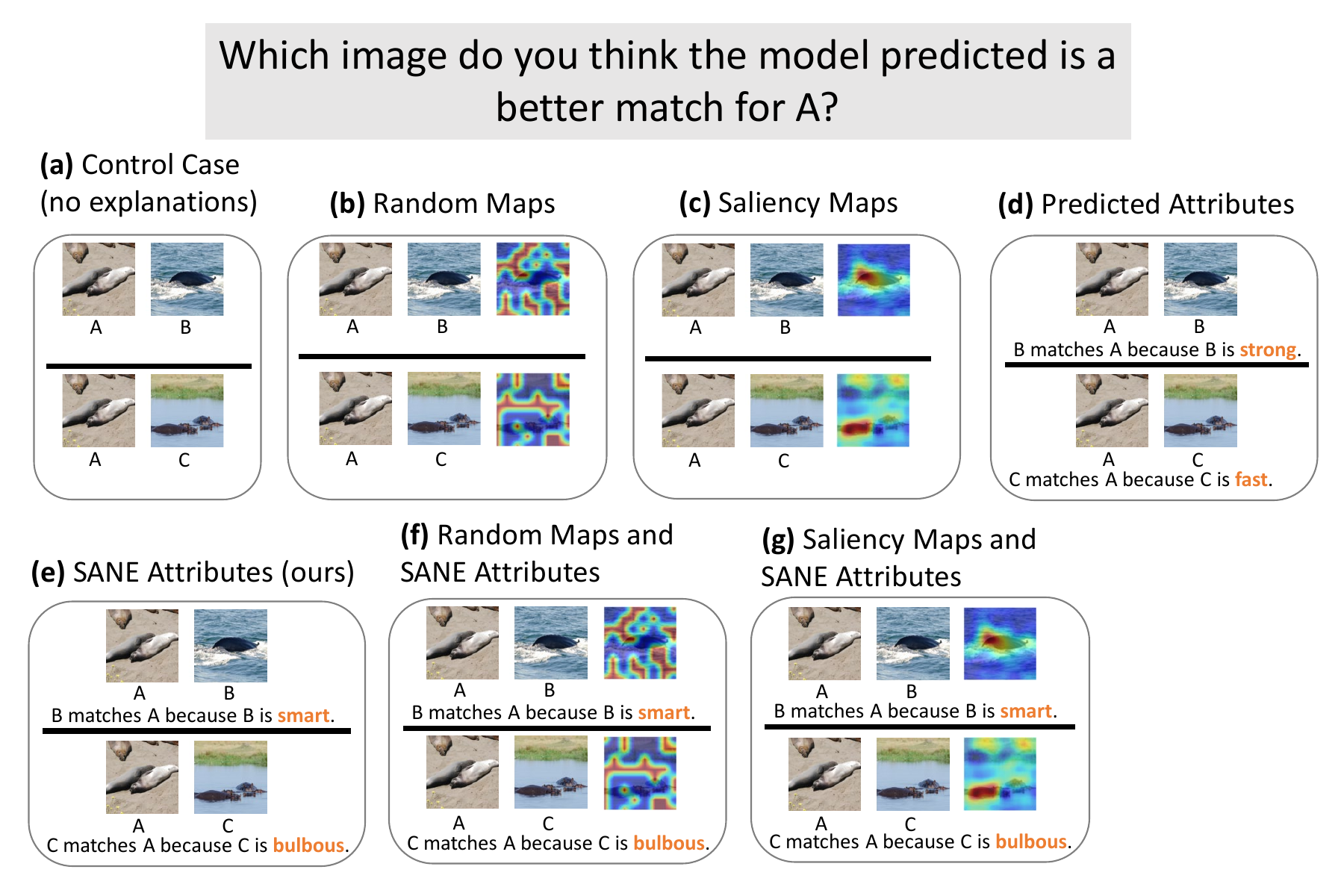}
    \caption{Variants of our user study for a given image triplet on the AwA dataset.}
    \label{fig:supp_user_study_1}
\end{figure}

An example of our user study variants for a given image triplet is shown in Figures~\ref{fig:supp_user_study} and~\ref{fig:supp_user_study_1}. The question asks users to select, given the image triplet $(A, B, C)$ presented, along with any additional information in each case, whether the image similarity model predicted $B$ or $C$ as a better match for $A$, as outlined in Section 4.3.

\newpage 
\section{Discovering Useful Attributes}
\label{sec:attr_discovered}

For datasets without attribute annotations, or those where the annotated attributes doesn't cover the extent of the visual attributes present in the dataset (\ie there are many unannotated attributes) we propose a method of discovering attributes that are useful for providing model explanations.  An attribute that is useful for explanations would commonly appear in the high importance regions of saliency maps.  When generating saliency maps for a query image, if many reference images attend to the same region of the query image then it is likely they are all matching to it for similar reasons (\ie there may be some attribute that they share which matches the query).  Given this observation, we discover attributes using the following saliency-based procedure:

\begin{enumerate}
    \item Obtain $K$ similar images for query image $q$ using k-NN.
    \item Generate a saliency map over $q$ for each of the similar (reference) images.
    \item Keep only those reference images which have their saliency peaks in the most common location (such as a unit square in a $7\times7$ grid) and pick top $N$ of them that have the highest similarity.
    \item For each reference image, generate its saliency map with $q$ and crop a $30 \times 30$ patch around the peak saliency region in the reference image.
    \item Upsample all the generated patches to full image resolution and get their embeddings.
    \item Cluster the patches produced for multiple queries $q$. Each cluster represents an attribute. If multiple patches were extracted from an image and they got assigned to different clusters, this image would be labeled with multiple attributes.
\end{enumerate}

\begin{figure}[ht]
  \centering
  \begin{subfigure}[b]{0.93\linewidth}
    \centering
    \includegraphics[width=\linewidth]{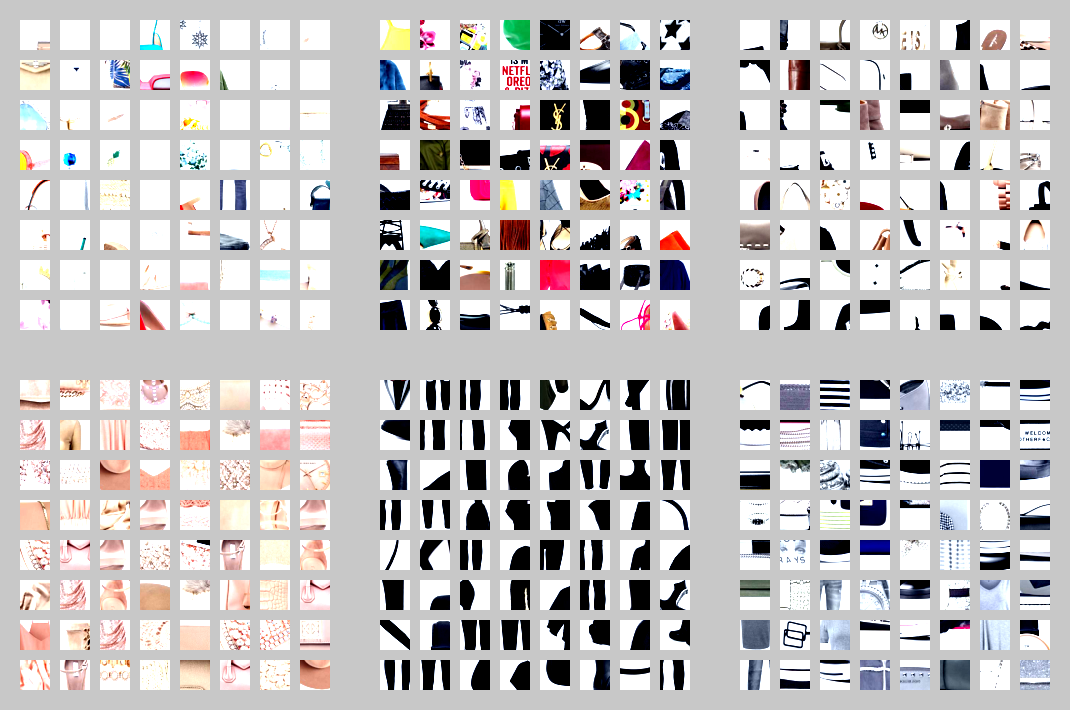}
    \caption{Patch-based clustering}
    \label{fig:clustering_patch}
  \end{subfigure}  
  
  \begin{subfigure}[b]{0.93\linewidth}
    \centering
    \includegraphics[width=\linewidth]{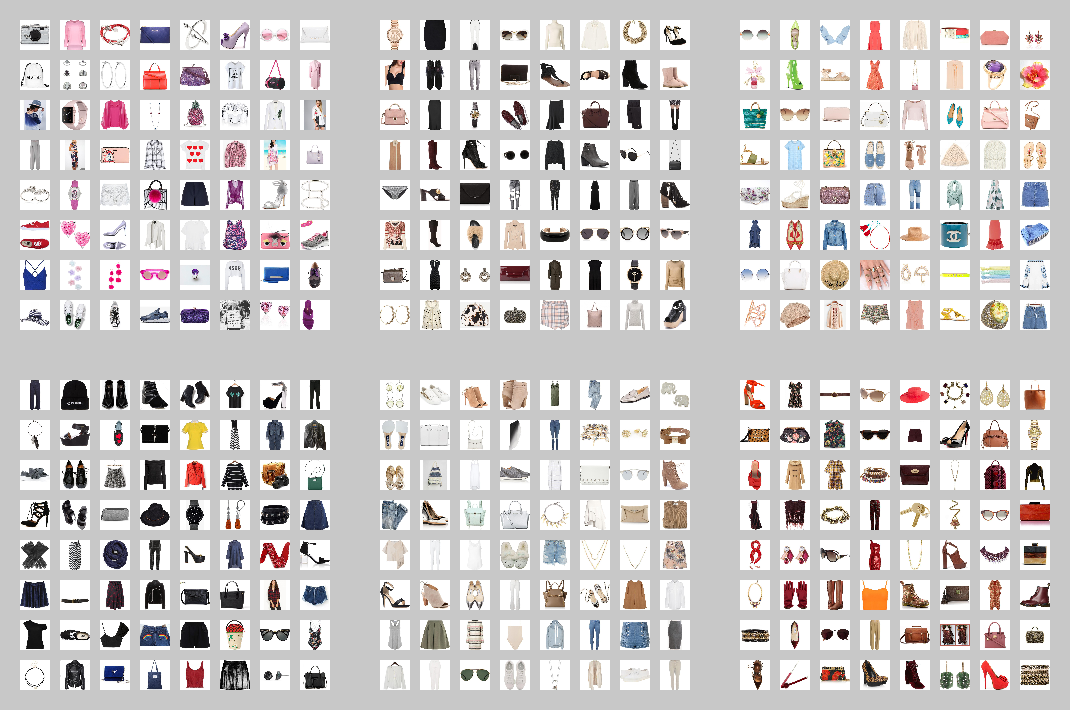}
    \caption{Full-frame clustering}
    \label{fig:clustering_full}
  \end{subfigure}

  \caption{Six clusters defining the attributes for two approaches to attribute discovery. (a) Each image is assigned a list of clusters that have patches from this image. Clustering is performed on salient patches. (b) Each image is assigned one of the clusters as an attribute. Clustering is performed on full-frame images.}
  \label{fig:clustering}
\end{figure}

Figure~\ref{fig:clustering_patch} illustrates the clustering produced by this procedure for a set of queries from Polyvore Outfits dataset. 

To evaluate this approach we compare it to randomly assigning images to clusters and to clustering based on their own embeddings, disregarding the saliency of image regions (Figure~\ref{fig:clustering_full}).  Saliency-based attribute discovery works best among the three unsupervised methods for Polyvore Outfits data, but full-frame clustering outperforms it for the AwA dataset (Table~\ref{tab:discovered_attribute}). We suspect the full frame clustering works better for AwA since it considers the background more than the patch-based method (Polyvore Outfits image's typically have white backgrounds).  In addition, our discovered attributes would likely be noisier due to the similarity model focusing on the background patches in some images as well.  Although our initial results are promising, attempting to discover attributes useful for explanations warrants additional investigation.

  


\begin{table}[t]
\setlength{\tabcolsep}{3pt}
    \centering
    \caption{Discovered attribute explanation performance comparison using the full SANE model.}
    \begin{tabular}{lcccccc|r}
    \hline
    & \multicolumn{2}{c}{Polyvore Outfits} & \multicolumn{2}{c}{Animals with Attributes 2} \\
    \hline
    Attribute Types & Insertion ($\uparrow$) & Deletion ($\downarrow$)  & Insertion ($\uparrow$) & Deletion ($\downarrow$)\\
    \hline
    \hline
    Random & 25.3 & -6.3 & 2.1 & -8.5\\
    Full Frame Discovery & 29.4 & -9.7 & 4.8 & -22.6\\ 
    Patch Discovery & 30.2 & -10.3 & 5.4 & -22.9\\
    Supervised Attributes & \textbf{31.5} & \textbf{-11.8} & \textbf{6.2} & \textbf{-24.1}\\
    \hline
    \end{tabular}
    \label{tab:discovered_attribute}
\end{table}
\end{document}